\documentclass[letterpaper, 10 pt, conference]{ieeeconf}  

\IEEEoverridecommandlockouts                              

\usepackage{amsmath,amssymb} 
\usepackage{color}
\usepackage{xcolor}
\usepackage{graphicx}
\usepackage{algorithm}
\usepackage{algorithmic}
\usepackage{subcaption}
\usepackage{adjustbox}
\usepackage{booktabs}
\usepackage{multirow}
\usepackage{threeparttable}
\usepackage{wrapfig}
\usepackage{caption}
\usepackage{pifont}
\newcommand{\cmark}{\ding{51}}%
\newcommand{\xmark}{\ding{55}}%

\newcommand{\ie}{\emph{i.e.}}
\newcommand{\eg}{\emph{e.g.}}

\title{\LARGE \bf
Cost-Aware Evaluation and Model Scaling for \\ LiDAR-Based 3D Object Detection
}

\author{Xiaofang Wang and Kris M. Kitani
\thanks{Both authors are with the Robotics Institute, Carnegie Mellon University. Email: \tt{\{xiaofan2,kkitani\}@cs.cmu.edu}.}
}

\begin{document}

\maketitle
\thispagestyle{empty}
\pagestyle{empty}

\begin{abstract}
Considerable research effort has been devoted to LiDAR-based 3D object detection and empirical performance has been significantly improved. While progress has been encouraging, we observe an overlooked issue: it is not yet common practice to compare different 3D detectors under the same cost, \eg, inference latency. This makes it difficult to quantify the true performance gain brought by recently proposed architecture designs. The goal of this work is to conduct a cost-aware evaluation of LiDAR-based 3D object detectors. Specifically, we focus on SECOND, a simple grid-based one-stage detector, and analyze its performance under different costs by scaling its original architecture. Then we compare the family of scaled SECOND with recent 3D detection methods, such as Voxel R-CNN and PV-RCNN++. The results are surprising. We find that, if allowed to use the same latency, SECOND can match the performance of PV-RCNN++, the current state-of-the-art method on the Waymo Open Dataset. Scaled SECOND also easily outperforms many recent 3D detection methods published during the past year. We recommend future research control the inference cost in their empirical comparison and include the family of scaled SECOND as a strong baseline when presenting novel 3D detection methods.
\end{abstract}


\section{Introduction}

LiDAR-based 3D object detection is essential for autonomous driving. Existing research efforts have proposed a diverse range of 3D detectors, where point clouds are organized in various formats (\eg, point-based~\cite{qi2018frustum,shi2019pointrcnn}, grid-based~\cite{yan2018second,lang2019pointpillars,yin2021center}, range view~\cite{bewley2020range,fan2021rangedet}, or hybrid~\cite{shi2020pv,shi2021pv,deng2021voxel} representation) and processed by different architecture components (\eg, PointNet~\cite{qi2017pointnet,qi2017pointnet++}, 3D sparse convolution~\cite{graham20183d}, or 2D convolution). These efforts result in a significant boost in detection performance. The Average precision (AP) for vehicle detection on the Waymo Open Dataset~\cite{waymo_open_dataset} has been improved from 56.62\% (PointPillars~\cite{lang2019pointpillars}) to 79.25\% (PV-RCNN++~\cite{shi2021pv}) in just three years!

However, despite the promising empirical performance, we make a somewhat worrisome observation on the current state of 3D detection research: the computational cost (\eg, inference latency) is usually not controlled during the comparison of different detectors. Recent 3D detection methods tend to emphasize and attribute the performance gain to their novel architecture design. But it is unclear whether the proposed detectors are faster or slower than the baselines they are comparing to.

Why are we worried about this observation? We note that when developing architectures on ImageNet~\cite{russakovsky2015imagenet}, it is common practice to compare them under the same cost~\cite{tan2019efficientnet,liu2019darts,radosavovic2020designing,liu2021swin,wang2021neighborhood,wang2022wisdom}. But this has yet to be the case for 3D object detection. Since simply scaling up an architecture can already boost the accuracy~\cite{tan2019efficientnet,bello2021revisiting}, it is unfair to compare different architectures without controlling the cost. Such an unfair comparison makes it unclear whether the performance gain in recent 3D detection methods is actually brought by their proposed architectural changes or simply due to the usage of more computation. This can cause misleading conclusions on the contribution of different architecture components.

By first addressing the unfair comparison problem, we can know the true contribution of the diverse architecture components used in existing methods. This is also important for future research to further push the frontier of 3D detection. Fully addressing this issue surely requires a community-wide effort. We take a step forward by analyzing the performance of a simple grid-based one-stage detector, \ie, SECOND~\cite{yan2018second}, under different costs by scaling its original architecture.

We choose SECOND for the following reasons: (1) SECOND is a widely-used baseline and generally believed to have been significantly outperformed; (2) SECOND has easy-to-use open-source implementation\footnote{We refer to the open-source implementation in OpenPCDet~\cite{openpcdet2020}.} available; and (3) most importantly, SECOND is the common part of several high-performing two-stage detectors (\eg, PV-RCNN++~\cite{shi2021pv} and Voxel R-CNN~\cite{deng2021voxel}), where SECOND is adopted as their first stage to generate region proposals. Studying the performance of SECOND can immediately inform us about whether these sophisticated second stage detectors are necessary to achieve competitive detection performance.

To analyze the performance of SECOND under different costs, we study how to scale its backbone. We show that increasing the pre-head resolution, \ie, the spatial dimension of the feature map being passed to the detection head, is often better than just increasing the network depth or width. A larger pre-head resolution allows a denser sampling of object anchors or keypoints.

Then we compare the family of scaled SECOND against recent 3D detection methods. The results are surprising. We find that, SECOND can easily outperform most recent 3D detection methods after being scaled up. Notably, scaled SECOND can match the performance of PV-RCNN++, the current state-of-the-art on the Waymo Open Dataset, if allowed to use a similar inference latency. Scaled SECOND also easily outperforms many recent methods published during the past year. Our results indicate that the gain brought by the architectural innovation in many recent methods is not as significant as what was shown in their papers.

We summarize our contributions as follows: (1) We point out the vast importance of comparing different 3D detectors under the same cost, which may sound obvious but was overlooked in the recent literature.
(2) We provide an extensive analysis on how to scale up the backbone of grid-based 3D detectors, \eg, SECOND, and find that increasing the pre-head resolution is a reliable source for better performance.
(3) We introduce the family of scaled SECOND by scaling up the original backbone of SECOND and conduct a cost-aware comparison of scaled SECOND against recent 3D detection methods. Our comparison leads to a surprising observation: simply scaling the backbone in SECOND can already match the state-of-the-art performance on the Waymo Open Dataset.

\section{Related Work}
\subsection{LiDAR-Based 3D Object Detection}

Point clouds captured by LiDAR sensor are irregular. This makes it difficult to directly apply traditional convolutional architectures to point clouds, which have been successful for images and videos but require the input data to be organized in the format of regular grids~\cite{li2021lidar}. Several different ways have been proposed to address this issue. Following the categorization of 3D detectors in PV-RCNN++~\cite{shi2021pv}, we briefly review existing 3D detection methods based on how they represent and process the point cloud.

\textbf{Point-based Representation.}
This line of work treats point clouds as unordered point sets and directly processes the raw point cloud. Most of them adopted PointNet or its variant~\cite{qi2017pointnet,qi2017pointnet++} as the backbone~\cite{qi2018frustum,shi2019pointrcnn,yang2019std,yang20203dssd,qi2019deep,qi2020imvotenet}.
Point-GNN~\cite{shi2020point} explored using graph neural networks to encode the point cloud by constructing a fixed radius near-neighbors graph. Pan et al.~\cite{pan20213d} proposed PointFormer, a Transformer architecture for 3D point clouds, to serve as the backbone in point-based detectors. Point-based representation can fully reserve the 3D structure and fine details. But the nearest neighbor search operation used in PointNet or PointNet++ variant is computationally prohibitive as the number of points increases. While the efficiency issue can be partially mitigated by downsampling the point cloud (\eg, only keeping 16384 points~\cite{shi2019pointrcnn}), the downsampling inevitably brings performance drop. This limits the application of point-based detectors to large-scale scenes~\cite{li2021lidar}.

\textbf{Grid-based Representation.}
To deal with the irregularity of point clouds, previous work proposed to divide point clouds into (1) regular grids, \eg, voxels, pillars, and (2) bird's-eye view (BEV), to make it possible to apply convolutional operations. VoxelNet~\cite{zhou2018voxelnet} partitioned the space into equally spaced voxels, applied PointNet~\cite{qi2017pointnet} in each voxel to generate voxel features, and then used dense 3D convolution to further aggregate the spatial context. SECOND~\cite{yan2018second} improved upon VoxleNet by using 3D sparse convolution~\cite{graham20183d} and removing the PointNet. PointPillars~\cite{lang2019pointpillars} proposed to organize the point cloud as pillars (vertical columns) to improve the voxel backbone efficiency. Voxel features or pillar features are often projected onto the ground plane, \ie, BEV, before being passed to the detection head, where 2D convolution can be readily applied.
Given BEV feature maps, CenterPoint~\cite{yin2021center} proposed a center-based detection head for 3D object detection without pre-defining axis-aligned anchors. The grid-based representation makes it easy to apply convolutional operations, but suffers from the quantization error caused by dividing the space into regular grids, which can limit the detection performance, especially for distant objects with a few points.

\textbf{Range View Representation.}
Range view is a commonly used representation of LiDAR data and can be efficiently processed by 2D convolutional architectures~\cite{meyer2019lasernet}. VeloFCN~\cite{li2016vehicle} is the pioneering work in this line, where they designed a fully convolutional network to detect 3D objects from range images. LaserNet~\cite{meyer2019lasernet} also used a fully convolutional network as the backbone. RCD~\cite{bewley2020range} proposed a novel range-conditioned dilation layer to account for the scale variation of objects in range images. RangeDet~\cite{fan2021rangedet} proposed several strategies to improve pure rage-view-based object detection. Challenges of using range view representation include dealing with scale variation and occlusion~\cite{meyer2019lasernet}.

\textbf{Hybrid Representation.}
Since different representations have their own pros and cons, previous methods~\cite{chen2017multi,ku2018joint,liang2018deep,zhou2019end,sun2021rsn,shi2020pv,shi2021pv,guan2022m3detr} also explored combining multiple representations of the point cloud. For example, MVF~\cite{chen2017multi} proposed a novel multi-view fusion algorithm to effectively use BEV and range view. M3DETR~\cite{guan2022m3detr} explored fusing multi-representation features from points, voxels and BEV with Transformer. Shi et al.~\cite{shi2020pv} proposed PV-RCNN, a two-stage detection framework that takes the advantage of both the voxel-based and point-based methods. Its extension PV-RCNN++~\cite{shi2021pv} achieved state-of-the-art performance on the Waymo Open Dataset.

\subsection{Characterizing Architectures}
ImageNet~\cite{russakovsky2015imagenet} has a large impact on designing novel architectures~\cite{krizhevsky2012imagenet,szegedy2015going,he2016deep,howard2017mobilenets,Zhang_2018_CVPR,tan2019efficientnet,dosovitskiy2021image} and is the de facto benchmark for evaluating novel architectures.
We note that when intending to show a novel architecture is more accurate than previous ones on ImageNet, the common practice is to ensure that the proposed architecture use similar computational cost with the baselines\footnote{Equivalently, it is also common in practice to show the proposed architecture can achieve comparable accuracy with less cost. e.g., FLOPs or latency.}~\cite{tan2019efficientnet,liu2019darts,radosavovic2020designing,liu2021swin,wang2021neighborhood,wang2022wisdom}. For example, RegNet~\cite{radosavovic2020designing} considered a wide range of computation regimes and conducted the comparison of architectures within each regime. The very recent Swin Transformer~\cite{liu2021swin} reported model parameters, FLOPs, and throughput in their evaluation.

Previous results~\cite{tan2019efficientnet,bello2021revisiting} have demonstrated that scaling up an architecture, \eg, increasing the number of layers or channels, can significantly improve the accuracy. Therefore, comparing architectures of different costs is unfair and cannot justify that the gain is due to the novel architecture design. Unfortunately, the aforementioned standard practice has not yet been ubiquitously adopted in 3D object detection. Our work is inspired by this standard practice and aims to quantify how much performance gain in state-of-the-art 3D detection methods is due to architectural innovation.

\section{Architecture Overview}
\label{sec:architecture-overview}

\begin{figure}[t]
\centering
\includegraphics[width=0.45\textwidth]{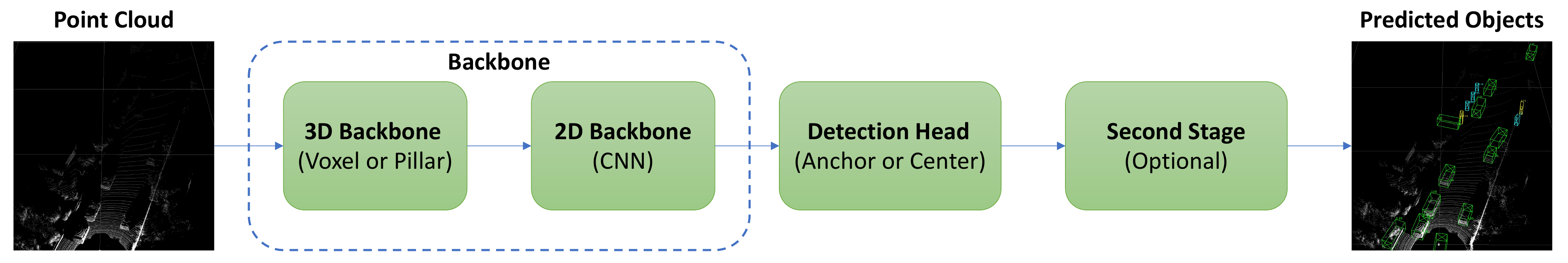}
\caption[Main components in a LiDAR-based 3D object detector.]{Main components in a LiDAR-based 3D object detector. We list example choices for a component in the bracket but there could be other choices.}
\label{fig:detector}
\end{figure}

\begin{table}[t]
\centering
\captionof{table}[Component Overview of Relevant Detectors.]{Component Overview of Relevant Detectors.}

\label{table:detector}
\adjustbox{width=0.49\textwidth}{
\begin{threeparttable}
\begin{tabular}{lccc}
\toprule
 & 3D Backbone & Detection Head  & Second Stage \\\midrule
\tnote{$\dagger$} SECOND-Anchor~\cite{yan2018second} & Voxel & Anchor & \xmark \\
\tnote{$\ddagger$} SECOND-Center & Voxel & Center & \xmark \\
PointPillars~\cite{lang2019pointpillars} & Pillar & Anchor & \xmark \\
CenterPoint~\cite{yin2021center} & Pillar / Voxel & Center & \cmark \\
PV-RCNN~\cite{shi2020pv} & Voxel & Anchor / Center & \cmark \\
PV-RCNN++~\cite{shi2021pv} & Voxel & Anchor / Center & \cmark
\\\bottomrule
\end{tabular}
\footnotesize{\begin{tablenotes}
\item[$\dagger$] SECOND-Anchor is the original SECOND method using anchor head.
\item[$\ddagger$] SECOND-Center is equivalent to the first stage of CenterPoint.
\end{tablenotes}}
\end{threeparttable}}
\end{table}

This section reviews the architecture details of relevant detection methods to provide background for our analysis. We illustrate a detector as the combination of a backbone, a detection head, and optionally a second stage in Fig.~\ref{fig:detector}. The backbone is further divided into a 3D one and a 2D one. Table~\ref{table:detector} gives a component overview of relevant detectors and we describe more details as follows.

\begin{figure}[t]
\centering
\includegraphics[width=0.45\textwidth]{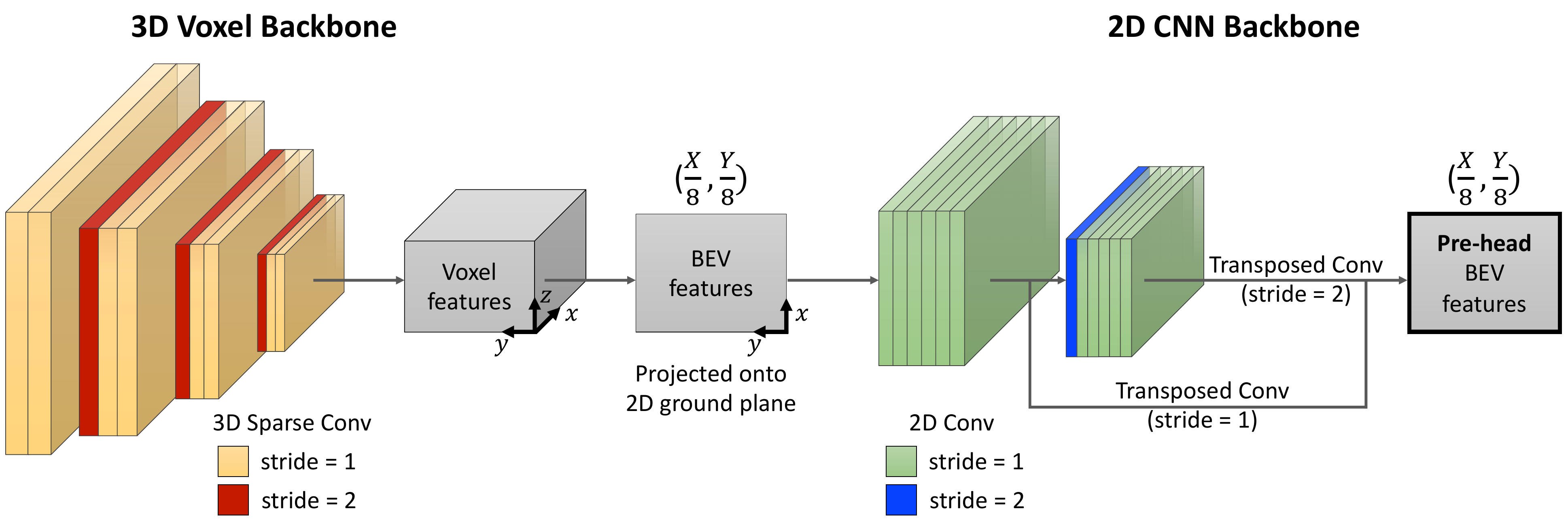}
\caption[Backbone Architecture of SECOND.]{Backbone Architecture of SECOND (A0 in Table~\ref{table:archtecture}). Assuming the input point cloud is initially grouped into $X\times Y \times Z$ voxels, the pre-head resolution of the shown A0 backbone, \ie, the spatial dimension of the obtained pre-head BEV features, will be $(\frac{X}{8}, \frac{Y}{8})$.}
\label{fig:backbone}
\end{figure}
\begin{table}[t]
\centering
\captionof{table}[List of architectures used in our analysis.]{List of architectures used in our analysis. A0 is the original backbone of SECOND implemented in OpenPCDet.}

\label{table:archtecture}
\adjustbox{width=0.49\textwidth}{
\begin{threeparttable}
\begin{tabular}{lcccccc}
\toprule
& 3D Depth & 3D Width & 2D Depth & 2D Width & Pre-Head \\\midrule
A0 & 2, 3, 3, 3 & 16, 32, 64, 64 & 6, 6 & 128, 256 & $(X / 8, Y / 8)$ \\
A0-deep & 8, 12, 12, 12 & 16, 32, 64, 64 & 24, 24 & 128, 256 & $(X / 8, Y / 8)$ \\
A0-wide & 2, 3, 3, 3 & 32, 64, 128, 128 & 6, 6 & 256, 512 & $(X / 8, Y / 8)$ \\
A0-d\&w & 3, 5, 5, 5 & 28, 56, 112, 112 & 9, 9 & 224, 448 & $(X / 8, Y / 8)$ \\\midrule
A1 & 2, 4 ,4 & 32, 64, 64 & 6, 6 & 128, 256 & $(X / 4, Y / 4)$ \\
A2 & 3, 6, 6 & 48, 96, 144 & 12, 12 & 128, 256 & $(X / 4, Y / 4)$
\\\bottomrule
\end{tabular}
\end{threeparttable}}
\end{table}

\subsection{SECOND, PointPillars \& CenterPoint}

\textbf{SECOND.}
Most of our analysis focuses on SECOND~\cite{yan2018second}, one of the earliest 3D detection methods and a widely-used baseline in the literature. SECOND first groups the point cloud into voxels and then extracts 3D voxel-wise features using the 3D backbone. The sparse 3D voxel-wise features are then projected onto the ground plane ($x$ and $y$-axis) to obtain dense 2D BEV features, which is done by channel concatenation across the height dimension ($z$-axis). The obtained BEV features are then processed by the 2D backbone and passed to the detection head to generate a set of region proposals, including their location, size, orientation and class. Finally, non-maximum suppression is applied on the region proposals to remove redundant object predictions.

Fig.~\ref{fig:backbone} shows the original backbone architecture of SECOND implemented in OpenPCDet~\cite{openpcdet2020} (A0 in Table~\ref{table:archtecture}). The 3D backbone is formed by stacking 3D sparse convolutional layers~\cite{graham20183d} and consists of four stages. The first layer at each stage, except the first stage, has a stride of 2 to reduce the spatial dimension in 3D. The 2D backbone consists of 2D convolution layers and has two stages. All the layers have a stride of 1 except that the first layer at the second stage has a stride of 2. The output of each stage is transformed or upsampled via transposed convolution and then concatenated across the channel dimension to obtain the final BEV features, which we will refer to as ``pre-head features'' since the features will be directly passed to the detection head to predict region proposals. Relatedly, the spatial dimension of the pre-head features given by A0 will be $(\frac{X}{8}, \frac{Y}{8})$, assuming the input point cloud is initially grouped into $X\times Y \times Z$ voxels. We discuss the details of detection head after PointPillars.

\textbf{PointPillars.} We also include PointPillars~\cite{lang2019pointpillars} in our analysis, another popular baseline for 3D detection. PointPillars~\cite{lang2019pointpillars} is similar to SECOND except for the 3D backbone, where the point cloud is organized as pillars (vertical columns) instead of voxels. They obtain pillar features by aggregating point features inside each pillar and then convert pillar features into a pseudo-image, \ie, BEV features, which are passed to the 2D backbone and detection head.

\textbf{Detection Head.}
As listed in Table~\ref{table:detector}, the original SECOND and PointPillars method use an anchor-based detection head, which pre-defines axis-aligned anchors of different classes on each location. We refer to the original SECOND method using anchor head as SECOND-Anchor. The anchor head takes BEV features as input and uses a convolutional layer to regress the residuals between the ground truth object boxes and pre-defined anchors, as well as predict the class probabilities of each anchor.

CenterPoint~\cite{yin2021center} proposed a center-based detection head, which dose not require pre-defining anchors and achieves superior performance over the anchor head. This center head first detects object centers using a keypoint detector and then regresses other attributes, \eg, object size and orientation, for each detected center. The center head is generic and can be used as a drop-in replacement for anchor head. Therefore, our analysis also considers the center head and uses it within SECOND by replacing the anchor head in SECOND-Anchor as center head, which we list as SECOND-Center in Table~\ref{table:detector}. For clarification, the full method of CenterPoint is a two-stage detector and SECOND-Center is exactly the same as the first stage of CenterPoint.

\subsection{Part-A2-Net, PV-RCNN, PV-RCNN++ \& Voxel R-CNN}
Part-A2-Net~\cite{shi2020points}, PV-RCNN~\cite{shi2020pv}, PV-RCNN++~\cite{shi2021pv}, and Voxel R-CNN~\cite{deng2021voxel} are all two-stage 3D detectors achieving competitive performance. They all use SECOND as their first stage to generate initial region proposals, which are then further refined in the second stage. Similar to SECOND, both anchor head and center head can be used in these two-stage detectors. Notably, PV-RCNN++ achieves state-of-the-art performance on the Waymo Open Dataset.

\section{Experimental Setup}
\label{sec:exp-setup}

We conduct experiments with OpenPCDet, an open-source code base for LiDAR-based 3D object detection. OpenPCDet is the official code release of Part-A2-Net~\cite{shi2020points}, PV-RCNN~\cite{shi2020pv}, PV-RCNN++~\cite{shi2021pv}, and Voxel R-CNN~\cite{deng2021voxel}, and also supports many other methods, including SECOND~\cite{yan2018second} and PointPillars~\cite{lang2019pointpillars}.

\textbf{Dataset and Metrics.}
We use the Waymo Open Dataset~\cite{waymo_open_dataset}, the largest public benchmark for LiDAR-based 3D object detection, in our experiments. It contains 798 train sequences ($\sim$158k frames) and 202 validation sequences ($\sim$40k frames). Following the standard protocol, we adopt average precision (AP) and average precision weighted by heading (APH) as the metrics and evaluate in two difficulty levels (LEVEL\_1 and LEVEL\_2).

\textbf{Training Setup.} By default, we train on all the train sequences and evaluate on all the validation sequences (100\% training setup). To save training time for the analysis in Sec.~\ref{sec:scale-analysis}, we adopt the 20\% training setup provided in OpenPCDet~\cite{openpcdet2020}. Under this setup, we train on 20\% frames uniformly sampled from the train sequences but still evaluate on all the validation sequences. Since LiDAR frames in one sequence are highly correlated, 20\% of data is usually representative enough.

\textbf{Inference Latency.} We use batch size 1 when measuring the latency of a detector, following the convention in detection~\cite{tan2020efficientdet}. The latency is measured on a Nvidia GeForce RTX 3090 GPU and a AMD EPYC 7402 24-Core CPU.

\section{Depth, Width, and Pre-Head Resolution}
\label{sec:scale-analysis}

Scaling the depth (number of layers), width (number of channels), or input image resolution have been widely used to improve the performance of a model~\cite{he2016deep,howard2017mobilenets,tan2019efficientnet,bello2021revisiting}. But it is still an open question about what the optimal scaling strategy is. There can be a large performance variation among different scaling configurations even when then the amount of cost is controlled, especially in the large computation regime~\cite{bello2021revisiting}. For example, EfficientNet~\cite{tan2019efficientnet} demonstrated that compound scaling, \ie, scaling all three dimensions including depth, width, and resolution, is better than scaling only one of the dimensions. Bello et al.~\cite{bello2021revisiting} observed diminishing returns in very large image resolutions in EfficientNet and suggested that one should increase the resolution slowly. To form the basis of our analysis, this section analyzes different ways to scale the backbone of SECOND.

\setlength{\tabcolsep}{2.5pt}
\begin{table}[t]
\centering
\caption[Performance of SECOND-Anchor with different backbones.]{Performance of SECOND-Anchor with different backbones on the Waymo validation set (20\% training). Scaling the pre-head resolution from $(\frac{X}{8}, \frac{Y}{8})$ to $(\frac{X}{4}, \frac{Y}{4})$ significantly improve the performance on all classes. But further increasing the resolution does not help.}

\label{table:anchor-resolution}
\adjustbox{width=0.45\textwidth}{
\begin{threeparttable}
\begin{tabular}{lcccccccccccccc}
\toprule
\multirow{2}{*}{Anchor Head} & \multicolumn{4}{c}{LEVEL\_2 3D APH} & Latency & Params & Memory & Pre-Head \\
 & Vehicle & Pedestrian & Cyclist & mAPH & (ms) & (M) & (GB) & Resolution \\\midrule
A0 & 62.02 & 47.49 & 53.53 & 54.35 & 27 & 5.33 & 5.8 & $(X / 8, Y / 8)$ \\
A0+Upsample & 64.61 & 51.92 & 59.81 & 58.78 & 38 & 5.69 & 9.0 & $(X / 4, Y / 4)$ \\
A0+Upsample$\times$2 & 64.51 & 50.50 & 59.82 & 58.28 & 52 & 5.69 & 20.0 & $(X / 2, Y / 2)$ \\
A1 & \textbf{65.65} & \textbf{59.20} & \textbf{64.33} & \textbf{63.06} & 65 & 5.56 & 14.4 & $(X / 4, Y / 4)$
\\\bottomrule
\end{tabular}
\end{threeparttable}}

\caption[Performance of SECOND-Center with different backbones.]{Performance of SECOND-Center with different backbones on the Waymo validation set (20\% training). The advantage of scaling pre-head resolution generalizes to the center-based detection head.}
\label{table:center-resolution}
\adjustbox{width=0.45\textwidth}{
\begin{threeparttable}
\begin{tabular}{lcccccccccccccc}
\toprule
\multirow{2}{*}{Center Head} & \multicolumn{4}{c}{LEVEL\_2 3D APH} & Latency & Params & Memory & Pre-Head \\
 & Vehicle & Pedestrian & Cyclist & mAPH & (ms) & (M) & (GB) & Resolution \\\midrule
A0 & 62.65 & 58.23 & 64.87 & 61.92 & 28 & 5.78 & 5.6 & $(X / 8, Y / 8)$ \\
A0+Upsample & 64.86 & 61.26 & 65.79 & 63.98 & 43 & 6.14 & 10.0 & $(X / 4, Y / 4)$ \\
A1 & \textbf{64.92} & \textbf{65.35} & \textbf{67.96} & \textbf{66.08} & 67 & 6.01 & 14.7 & $(X / 4, Y / 4)$
\\\bottomrule
\end{tabular}
\end{threeparttable}}
\vspace{-2.0em}
\end{table}

\subsection{Pre-Head Resolution}

Unlike images, there is no notion of resolution for raw point clouds. Therefore, we consider the pre-head resolution, \ie, the spatial dimension of the pre-head BEV features, as an alternative choice to scale the backbone.

We note that previous work on 2D object detection~\cite{lin2017feature,ghiasi2019fpn,tan2020efficientdet} explored using multi-scale feature maps, whose main motivation is to handle the scale variation of objects in images, \ie, using a higher-resolution feature map for larger anchors, since the size of the same object could vastly vary depending on its distance to the camera. But this is not the case for LiDAR point clouds as the size of a specific object in point clouds is fixed no matter its distance to the sensor.

The advantage of a larger pre-head resolution for LiDAR-based detection is in the denser sampling of anchors or keypoints. The anchor head places pre-defined anchors on every location of the BEV features. Therefore, a larger pre-head feature map resolution means more anchors are used. For center head~\cite{yin2021center}, which detects object centers via keypoint estimation, a larger resolution means more keypoints are classified.

Now we empirically investigate whether scaling the pre-head resolution leads to performance improvement. Starting from A0, the original backbone in SECOND with a pre-head resolution of $(\frac{X}{8}, \frac{Y}{8})$, we consider the following backbones to increase the pre-head resolution:
\begin{itemize}
\item \textbf{A0+Upsample:} we change the stride of the transposed convolution at the end of first stage in the 2D backbone to 2, and add another transposed convolutional layer of a stride 2 at the end of the second stage. This yields a pre-head resolution of $(\frac{X}{4}, \frac{Y}{4})$.
\item \textbf{A0+Upsample$\times$2:} we further upsamples the pre-head feature map given by `A0+Upsample' by 2x with Bilinear interpolation. This yields a resolution of $(\frac{X}{2}, \frac{Y}{2})$.

\item \textbf{A1:} we remove the last stage in the 3D backbone of A0 to perform less downsampling in the network. The number of layers and channels are slightly adjusted so that A1 has a similar number of parameters to A0. A1 has a pre-head resolution of $(\frac{X}{4}, \frac{Y}{4})$.
\end{itemize}

We show the performance of different backbones in Table~\ref{table:anchor-resolution}. We also report the inference latency, number of parameters, and memory footprint at batch size 2 during training for completeness. Here we focus on analyzing whether increasing the pre-head resolution can improve the performance, so we do not control the latency of different backbones to be the same.

As shown in Table~\ref{table:anchor-resolution}, scaling the resolution to $(\frac{X}{4}, \frac{Y}{4})$ is beneficial as A1 and `A0+Upsample' outperform A0 by a large margin on all classes. Table~\ref{table:center-resolution} provides the results of SECOND-Center with different backbones and we see the benefit of a larger pre-head resolution generalizes to center head. But we also notice that further increasing the pre-head resolution does not bring any additional gain (`A0+Upsample' vs. `A0+Upsample$\times$2'). Therefore, we conclude that scaling the pre-head resolution is a reliable source for better performance but one should refrain from increasing the resolution aggressively.

\subsection{Depth vs. Width vs. Pre-Head Resolution}

\setlength{\tabcolsep}{2.5pt}
\begin{table}[t]
\centering
\caption[Depth vs. Width vs. Pre-Head Resolution.]{Depth vs. Width vs. Pre-Head Resolution. We compare SECOND-Anchor with different scaled backbones under the same latency on the Waymo validation set (20\% training). Scaling the pre-head resolution provides the highest overall mAPH with the fewest parameters.}

\label{table:depth-width-res-second}
\adjustbox{width=0.45\textwidth}{
\begin{threeparttable}
\begin{tabular}{lcccccccccccccc}
\toprule
\multirow{2}{*}{Anchor Head} & \multicolumn{4}{c}{LEVEL\_2 3D APH} & Latency & Params & Memory \\
 & Vehicle & Pedestrian & Cyclist & mAPH & (ms) & (M) & (GB) \\\midrule
A0 & 62.02 & 47.49 & 53.53 & 54.35 & 27 & 5.33 & 5.8 \\
\midrule
\multicolumn{4}{l}{\textbf{Without Residual Connections}}  &  &  \\
A0-deep & 59.98 & 35.53 & 43.15 & 46.22 & 55 & 20.92 & 13.5 \\
A0-wide & 65.10 & 53.01 & 59.61 & 59.24 & 61 & 19.96 & 8.0 \\
A0-d\&w & 65.50 & 51.94 & 59.46 & 58.97 & 68 & 23.76 & 9.9 \\
A1 & \textbf{65.65} & \textbf{59.20} & \textbf{64.33} & \textbf{63.06} & 65 & 5.56 & 14.4 \\\midrule
\multicolumn{4}{l}{\textbf{With Residual Connections}}  &  &  \\
A0-deep$_\text{res}$ & \textbf{66.98} & 54.80 & 60.35 & 60.71 & 59 & 21.23 & 15.4 \\
A0-wide$_\text{res}$ & 65.82 & 55.17 & 60.41 & 60.47 & 64 & 20.19 & 8.0 \\
A0-d\&w$_\text{res}$ & 66.69 & 56.00 & 61.70 & 61.46 & 66 & 21.66 & 10.4 \\
A1$_\text{res}$ & 65.78 & \textbf{59.82} & \textbf{64.27} & \textbf{63.29} & 67 & 5.65 & 13.7
\\\bottomrule
\end{tabular}
\end{threeparttable}}
\vspace{-2.0em}
\end{table}

We now compare the following ways to scale the backbone in SECOND: (1) depth only (A0-deep), (2) width only (A0-wide), (3) depth and width at the same time (A0-d\&w), and (4) pre-head resolution only (A1). The architecture details are available in Table~\ref{table:archtecture} and the results are shown in Table~\ref{table:depth-width-res-second}. We control the inference latency of of different backbones are to be similar for fair comparison.

While other backbones can easily improve the performance, we notice that A0-deep underperforms the original A0 backbone. We conjecture this is due to the lack of residual connections. A0-deep is much deeper than other networks and harder to train. Therefore, we add residual connections to all the backbones (subscripted by `res' in Table~\ref{table:depth-width-res-second}).

All the backbones benefit from adding residual connections and significantly outperform A0. Among the different choices, the compound scaling of depth and width is better than scaling one dimension only, echoing Tan et al.~\cite{tan2019efficientnet}. Scaling the pre-head resolution achieves the best overall performance, while scaling depth is the best for vehicle detection. Scaling the pre-head resolution also significantly saves the number of parameters compared with other choices.

\section{Scaled SECOND vs. Recent Methods}
\label{sec:fair-comparison}

\subsection{Family of Scaled SECOND}
Based on the above analysis on how to scale the backbone in SECOND, we introduce the family of scaled SECOND. In addition to the A0 and A1$_\text{res}$ backbone, we design a larger backbone A2$_\text{res}$ to cover the high latency regime. A2$_\text{res}$ is obtained by adding residual connections in A2, where A2 is obtained by increasing the depth and width of A1 (see Table~\ref{table:archtecture} for architecture details).

Then the family of scaled SECOND includes three backbones: A0, A1$_\text{res}$, and A2$_\text{res}$. As mentioned in Sec.~\ref{sec:architecture-overview}, both the anchor head and center head can be used within SECOND. So we have both SECOND-Anchor and SECOND-Center with the three backbones. To be clear, SECOND-Anchor is the original method of SECOND~\cite{yan2018second} and SECOND-Center replaces the anchor head in the original SECOND with the center head proposed in CenterPoint~\cite{yin2021center}.

Next, we will compare the family of scaled SECOND against recent LiDAR-based 3D object detection methods. We first compare scaled SECOND against several two-stage detectors in Sec.~\ref{sec:comparison-two-stage}. These two-stage detectors are specifically selected as they use SECOND as their first stage, including Part-A2-Net~\cite{shi2020points}, PV-RCNN~\cite{shi2020pv}, PV-RCNN++~\cite{shi2021pv}, and Voxel R-CNN~\cite{deng2021voxel}. Then we extend the comparison to more methods in Sec.~\ref{sec:full-comparison}.

\begin{figure*}[t]
\centering
\begin{subtable}[t]{0.21\textwidth}
\centering
\includegraphics[width=1.0\textwidth]{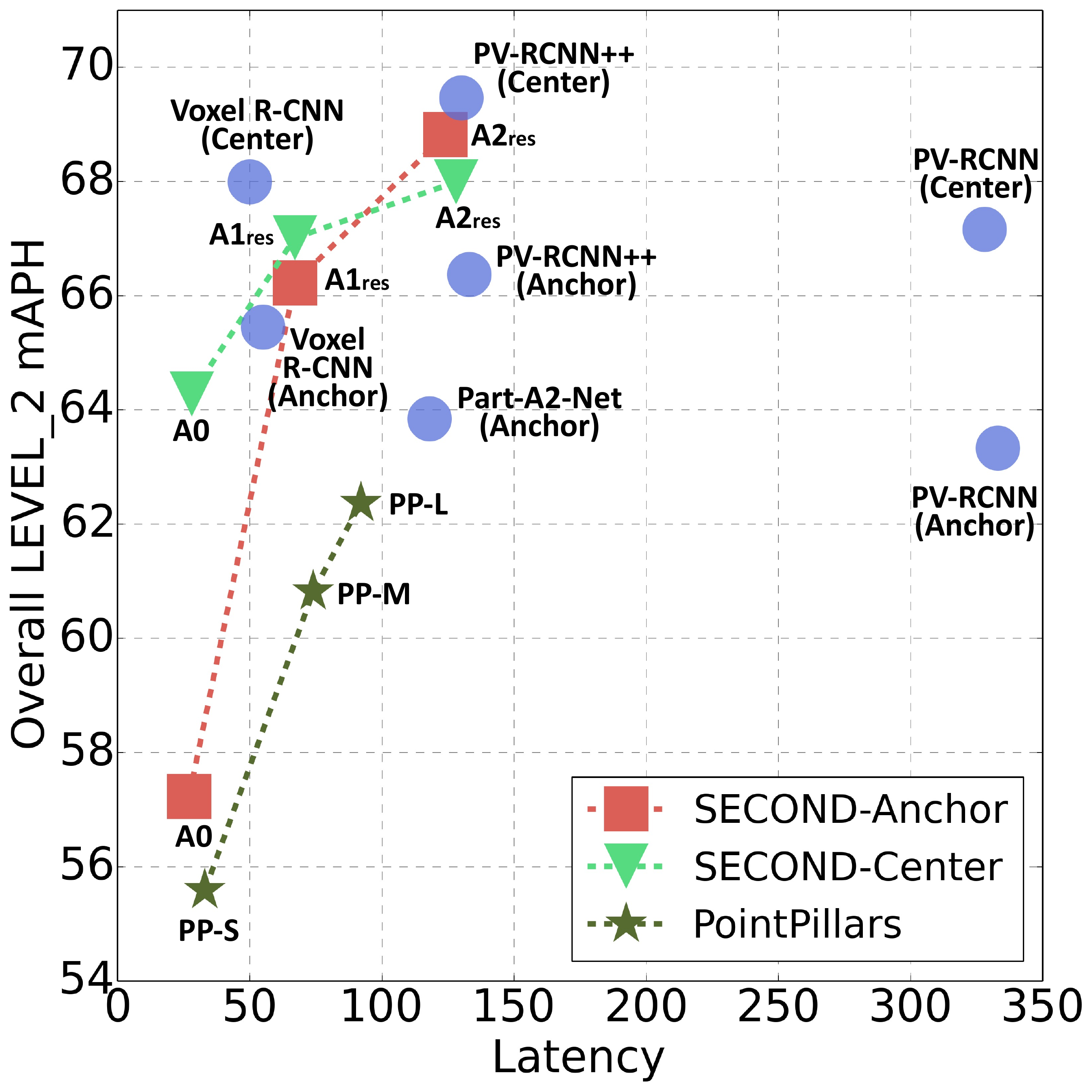}
\caption{Overall mAPH}
\label{fig:comparison-overall}
\end{subtable}
\begin{subtable}[t]{0.21\textwidth}
\centering
\includegraphics[width=1.0\textwidth]{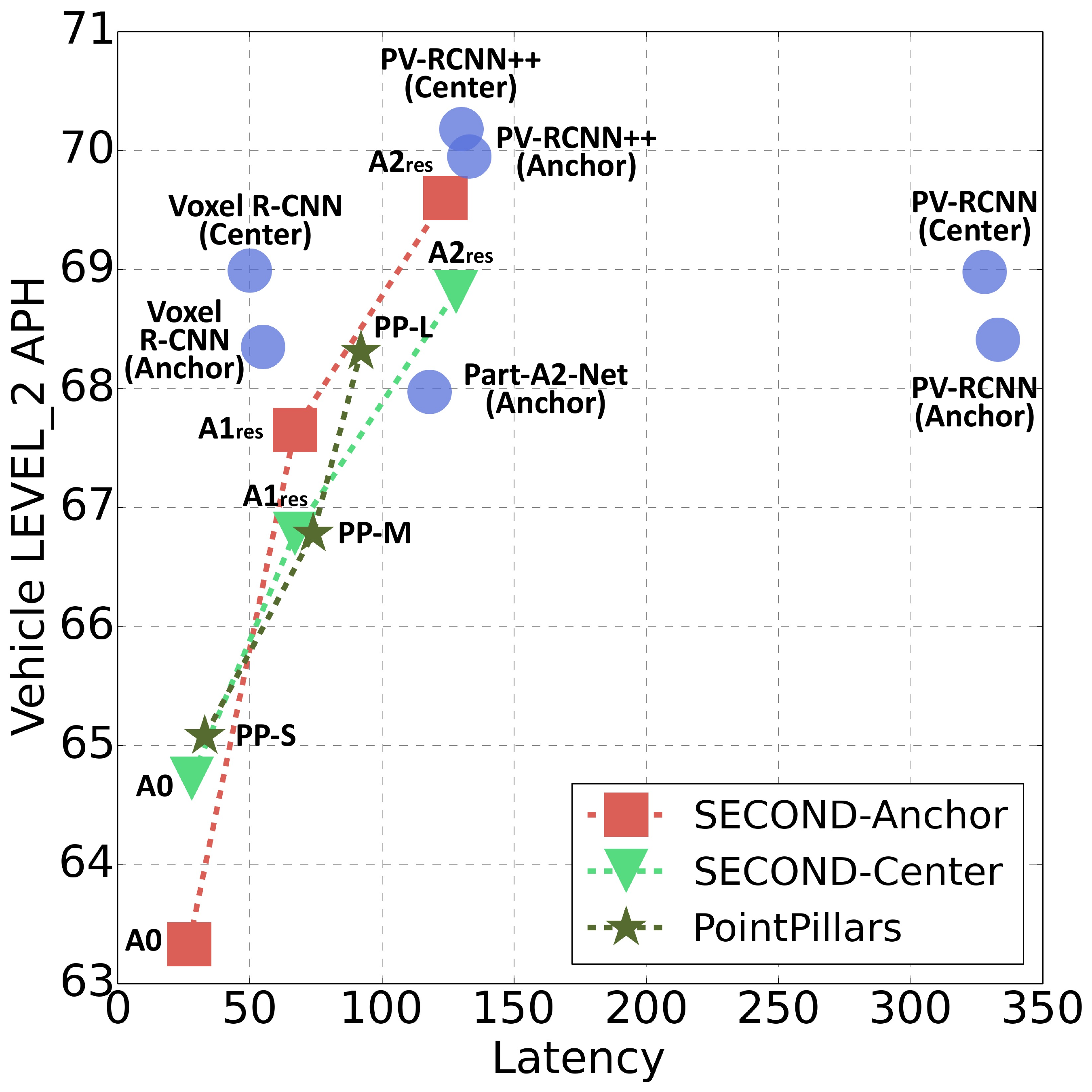}
\caption{Vehicle APH}
\label{fig:comparison-vehicle}
\end{subtable}
\begin{subtable}[t]{0.21\textwidth}
\centering
\includegraphics[width=1.0\textwidth]{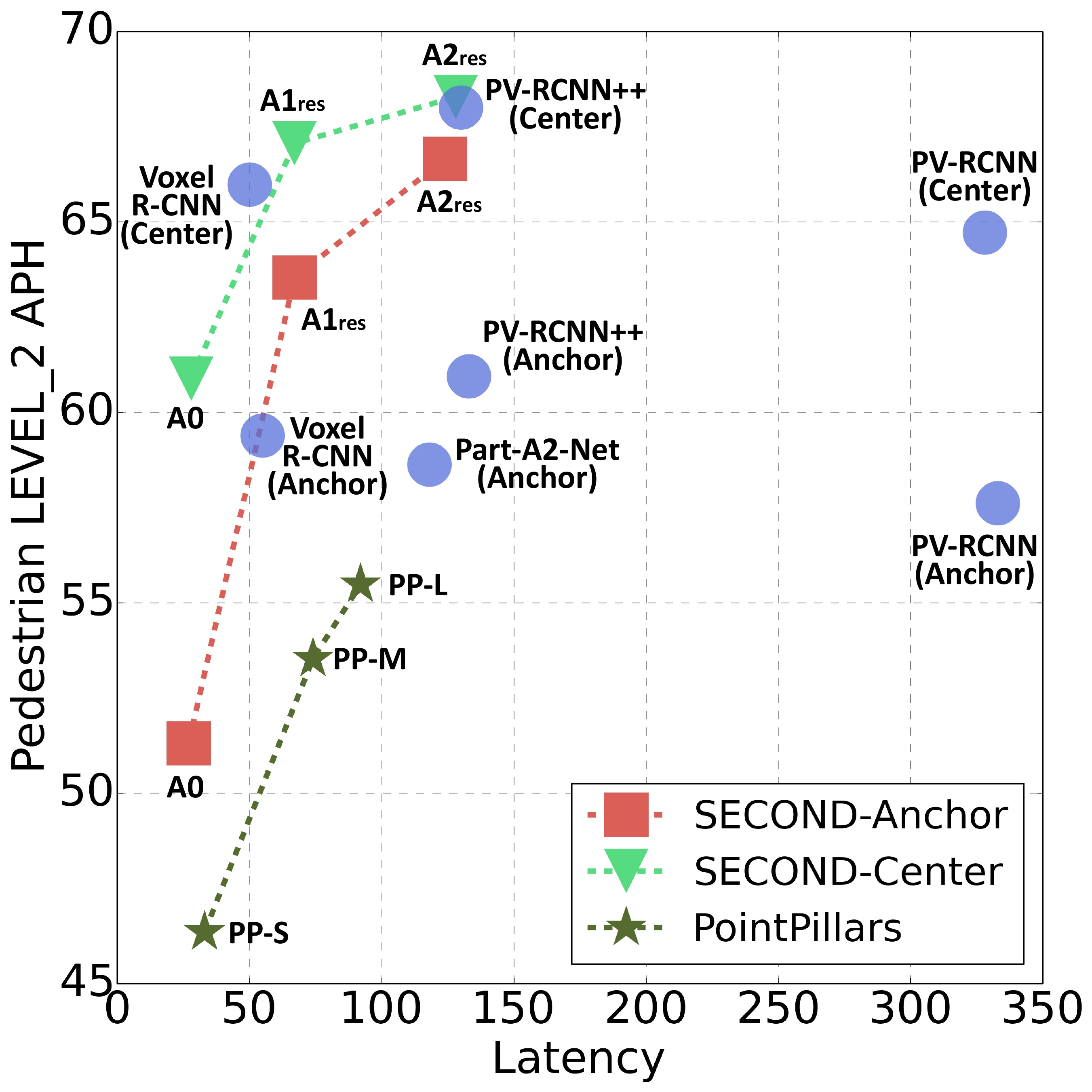}
\caption{Pedestrian APH}
\label{fig:comparison-pedestrian}
\end{subtable}
\begin{subtable}[t]{0.21\textwidth}
\centering
\includegraphics[width=1.0\textwidth]{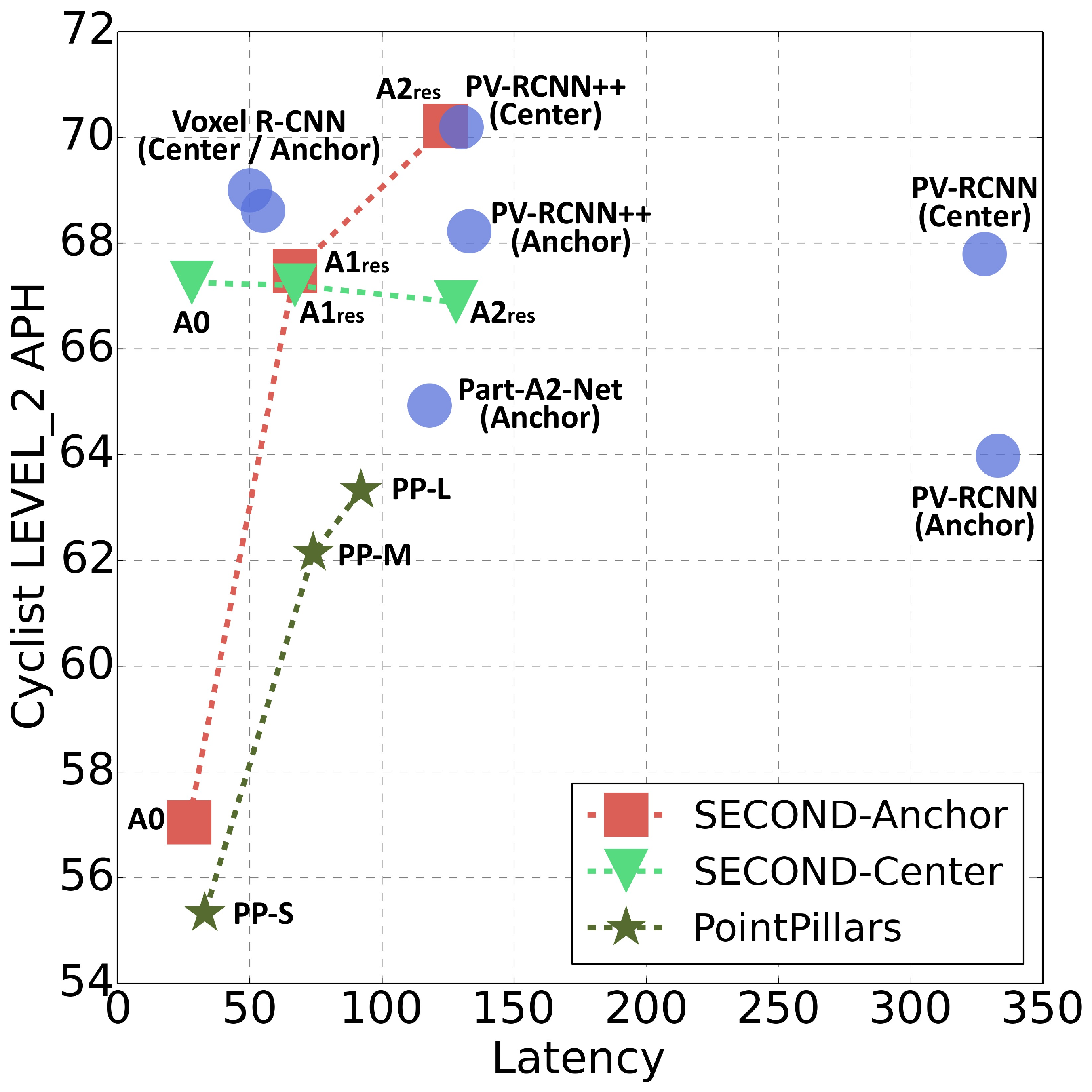}
\caption{Cyclist APH}
\label{fig:comparison-cyclist}
\end{subtable}

\caption[Scaled SECOND vs. Selected Two-Stage Detectors.]{Scaled SECOND vs. Selected Two-Stage Detectors. Only Voxel-RCNN (Center) and PV-RCNN++ (Center) can outperform the overall mAPH of scaled SECOND with a similar or smaller latency.}
\label{fig:comparison-two-stage}
\vspace{-1.5em}
\end{figure*}

\subsection{Comparison Against Selected Two-Stage Detectors}
\label{sec:comparison-two-stage}

We compare the family of scaled SECOND against the following two-stage detectors in Figure~\ref{fig:comparison-two-stage}: Part-A2-Net~\cite{shi2020points}, PV-RCNN~\cite{shi2020pv}, PV-RCNN++~\cite{shi2021pv}, and Voxel R-CNN~\cite{deng2021voxel}. As mentioned above, we select them as they all use SECOND as their first stage. We summarize the results in Figure~\ref{fig:comparison-two-stage}. All the results in Figure~\ref{fig:comparison-two-stage} are obtained using OpenPCDet~\cite{openpcdet2020}, the official implementation of the selected two-stage detectors, to ensure reproducibility and fair comparison. Since the detection head can have a big influence on the performance, we consider both the anchor head and center head for PV-RCNN, PV-RCNN++ and Voxel R-CNN in Figure~\ref{fig:comparison-two-stage} to provide a more complete comparison.

\textbf{Overall Comparison.} We observe from Figure~\ref{fig:comparison-overall} that scaled SECOND significantly outperforms Part-A2-Net and PV-RCNN after controlling the latency in the comparison. Only Voxel R-CNN and PV-RCNN++ can pass the test of scaled SECOND, \ie, more accurate than scaled SECOND when using a similar or smaller latency. But this only happens if they use the center head. Voxel R-CNN (Anchor) and PV-RCNN++ (Anchor) do not pass the test of scaled SECOND either.

\textbf{PV-RCNN++.} We take a closer look at PV-RCNN++ as it is the current state-of-the-art on the Waymo Open Dataset. The original SECOND method (SECOND-Anchor-A0) significantly underperforms PV-RCNN++ (Anchor). But simply scaling up its original backbone (A0) to A1$_\text{res}$ can easily achieve a similar mAPH with PV-RCNN++ (Anchor) while being twice faster during inference. The highest-performing method PV-RCNN++ (Center) is only slightly better than SECOND-Anchor-A2$_\text{res}$ if SECOND is allowed to use a similar latency. The performance gain brought by PV-RCNN++ is much smaller than what was shown in paper.

\textbf{Voxel R-CNN.} Voxel R-CNN (Center) outperforms SECOND-Center-A1$_\text{res}$ in the overall mAP with a smaller latency. The strong performance and efficiency of Voxel R-CNN is mainly due to their proposed Voxel RoI pooling~\cite{deng2021voxel} in the second stage, which does not use the expensive ball query to find nearest neighbors in the 3D space but rather uses an efficient voxel query operation.

\setlength{\tabcolsep}{2.5pt}
\begin{table}[t]
\centering
\caption{Scaled SECOND vs. Other Recent 3D Detection Methods. Most methods fail to outperform SECOND-Anchor-A1$_\text{res}$, which is exactly the same as the original SECOND~\cite{yan2018second} except for using a larger backbone.}
\label{table:comparison-other}
\adjustbox{width=0.49\textwidth}{
\begin{threeparttable}
\begin{tabular}{lccccccccccccccc}
\toprule
 &  & Latency & \multicolumn{2}{c}{Vehicle} & \multicolumn{2}{c}{Pedestrian} & \multicolumn{2}{c}{Cyclist} & Overall \\
 & Venue & (ms) & AP & APH & AP & APH & AP & APH & mAPH \\\midrule
PPBA~\cite{cheng2020improving} & ECCV 2020 & - & - & 53.4 & - & 53.9 & - & - & - \\
LiDAR R-CNN~\cite{li2021lidar} & CVPR 2021 & - & 64.7 & 64.2 & 63.1 & 51.7 & 66.1 & 64.4 & 60.10 \\
3D-MAN~\cite{yang20213d} & CVPR 2021 & - & 67.61 & 67.14 & 62.58 & 59.04 & - & - & - \\
PPC~\cite{chai2021point} & CVPR 2021 & - & - & 56.7 & - & 61.5 & - & - & - \\
MGAF-3DSSD~\cite{li2021anchor} & ACM MM 2021 & \tnote{$\dagger$} $\sim 60$ & 65.35 & - & - & - & - & - & - \\
RangeDet~\cite{fan2021rangedet} & ICCV 2021 &  \tnote{$\ddagger$} $\sim 58$ & 64.03 & 63.57 & 67.60 & \textbf{63.89} & 63.33 & 62.08 & 63.18 \\
VoTr-TSD~\cite{mao2021voxel} & ICCV 2021 & \tnote{$\sharp$} $>$ 300 & 65.91 & 65.29 & - & - & - & - & - \\
Pyramid-PV~\cite{mao2021pyramid} & ICCV 2021 & \tnote{$\sharp$} $>$ 300 & 67.23 & 66.68 & - & - & - & - & - \\
\textbf{SECOND-Anchor-A1$_\text{res}$} &  & 67 & \textbf{68.13} & \textbf{67.65} & \textbf{71.57} & 63.54 & \textbf{68.53} & \textbf{67.47} & \textbf{66.22} \\\midrule
Voxel-to-Point~\cite{li2021voxel} & ACM MM 2021 & - & 69.77 & - & - & - & - & - & - \\
\textbf{SECOND-Anchor-A2$_\text{res}$} &  & 124 & \textbf{70.06} & \textbf{69.60} & \textbf{73.98} & \textbf{66.65} & \textbf{71.22} & \textbf{70.21} & \textbf{68.82}
\\\bottomrule
\end{tabular}
\footnotesize{\begin{tablenotes}
\item[$\dagger$] We estimate its latency to be $\sim 60$ ms on Waymo as it is  $\sim$2x faster than Part-A2-Net on KITTI~\cite{li2021anchor}
\item[$\ddagger$] Measured as 12 fps on 2080Ti~\cite{fan2021rangedet}. We estimate its latency to be 58 ms on RTX 3090 based on~\cite{li2022gpubench}.
\item[$\sharp$] We estimate its latency to be larger than 300 ms as it is slower than PV-RCNN on KITTI~\cite{mao2021voxel,mao2021pyramid}.
\end{tablenotes}}
\end{threeparttable}}
\vspace{-2.0em}
\end{table}

\subsection{Full Comparison Against Recent Methods}
\label{sec:full-comparison}
We now extend the comparison to other recent 3D object detection methods in Table~\ref{table:comparison-other}. While all these methods are proposed after SECOND~\cite{yan2018second}, we observe that most methods fail to outperform SECOND-Anchor-A1$_\text{res}$, which is exactly the same as the original SECOND except for using a larger backbone. SECOND-Anchor-A1$_\text{res}$ is both faster and more accurate than some recent methods, such as VoTR-TSD~\cite{mao2021voxel} and Pyramid-PV~\cite{mao2021pyramid}.

The strong performance and simplicity of scaled SECOND should motivate future research to include them as baselines whenever proposing novel 3D object detectors.

\subsection{Comparison Under Multiple Latencies}
In the above, we only consider one specific latency when comparing two methods. The family of scaled SECOND contains backbones of different sizes and allows us to have a more complete comparison between methods under multiple latencies. We observe that the conclusion under one latency may not generalize to another and the ranking of two methods could flip.

\textbf{Anchor Head vs. Center Head.}
The center head was proposed in CeterPoint~\cite{yin2021center} and demonstrated impressive performance gain over the anchor head. But CenterPoint only experimented with a small backbone (similar size to A0). We observe that the benefit of center head shrinks as the backbone size grows. As shown in Figure~\ref{fig:comparison-overall}, the anchor head considerably underperforms the mAPH of center head when using A0. But after scaling up A0 to A2$_\text{res}$, the anchor head obtains a higher mAPH than the center head. We observe a similar trend for vehicle or pedestrian  detection. For example, for vehicle detection in Figure~\ref{fig:comparison-vehicle}, the ranking of anchor head and center head flips after the backbone is scaled up to A1$_\text{res}$.

\textbf{SECOND-Anchor vs. PointPillars.}
We observe that the ranking of SECOND-Anchor and PointPillars changes under different latencies. For example, PointPillars outperforms SECOND by $\sim 2\%$ on vehicle detection under the low latency regime (PointPillars-S vs. SECOND-Anchor-A0 in Figure~\ref{fig:comparison-vehicle}). The low latency regime is where the PointPillars focused on when it was proposed. However, as the backbone size grows, PointPillars-M underperforms SECOND-Anchor-A1$_\text{res}$ on vehicle detection.

\section{Conclusion}
To correctly evaluate the architecture design space of 3D detectors, we point out that it is important to compare different architectures under the same cost. Following this philosophy, we conduct an analysis of how to scale the backbone of SECOND, a simple baseline that is generally believed to have been significantly surpassed, and then introduce the family of scaled SECOND. Scaled SECOND sets a strong baseline for future research on 3D object detection: it outperforms most recent methods and can match the performance of the state-of-the-art method PV-RCNN++ on the Waymo Open Dataset if allowed to use a similar latency. We hope our analysis can encourage future research to adopt the good practice of cost-aware evaluation and include the family of scaled SECOND as a strong baseline when presenting novel 3D detection methods. We also show that the ranking of two methods can flip under different latencies and suggests that one should conduct the comparison under multiple latencies if possible.

\bibliographystyle{IEEEtranS}
\bibliography{egbib}

\end{document}